\documentclass[11pt]{article}

\usepackage[a4paper,margin=1in]{geometry}
\usepackage{amsmath,amssymb}
\usepackage{graphicx}
\usepackage{booktabs}
\usepackage{siunitx}
\usepackage{microtype}
\usepackage[numbers,sort&compress]{natbib}
\usepackage[hidelinks]{hyperref}
\hypersetup{
  pdfauthor={Phillip Jiang},
  pdftitle={When Does Advection-Aware Graph Nowcasting Help?}
}
\usepackage{xcolor}
\usepackage{caption}
\newcommand{\vhat}{\hat{\mathbf{v}}}
\newcommand{\kt}{k_t}

\title{When Does Advection-Aware Graph Nowcasting Help?\\
A Controlled Study of Distributed Solar Ramp Forecasting\\
with a Self-Supervised Cloud-Motion Estimator}

\author{%
  Phillip Jiang \\
  Appsofa LLC \\
  \texttt{phillip.jiang@appsofa.com}%
}
\date{\today}

\begin{document}
\maketitle

\begin{abstract}
Short-term forecasting of cloud-induced power ramps across a network of
distributed photovoltaic (PV) or irradiance sensors is a recognised pain point
for grid operators. A natural idea is to make the graph neural network (GNN)
\emph{advection-aware}: connect each site to the sites upwind of it, with edge
time-lags set by the cloud-motion vector (CMV), so that a ramp is propagated
forward before it physically arrives. Using a controlled synthetic testbed with a
known wind field, we show that (i) with a realistic cross-correlation CMV
estimate, an explicit advection graph does \emph{not} beat a plain static or
learned-adjacency spatiotemporal GNN; (ii) roughly half of the benefit available
from a perfect CMV comes simply from providing an accurate motion vector as an
input feature, not from graph structure; and (iii) advection helps only when the
advective displacement over the forecast horizon, $v\cdot H$, fits inside the
sensor network. Motivated by (ii), we introduce a small \emph{self-supervised}
cloud-motion estimator---a position-aware encoder trained only on a multi-lag
optical-flow reconstruction objective with an annealed kernel---that recovers the
true wind vector to \SI{2}{\degree}--\SI{4}{\degree} median angular error,
$2$--$4\times$ better than the classical cross-correlation method across every
wind regime. Freezing this estimator and feeding its vector to the forecaster
closes about \SI{60}{\percent} of the oracle-CMV RMSE gap at moderate wind
(\SIrange{8}{15}{\percent} RMSE reduction over no advection), with no external
wind data. We also report a negative result for a spatially-coherent
probabilistic head. All claims are established on a single synthetic simulator;
we discuss why real-network validation is the necessary next step and outline it.
\end{abstract}

\section{Introduction}
Integrating large amounts of solar generation exposes grid operators to
\emph{ramps}: sudden minutes-scale drops or surges in aggregate PV output as
cloud shadows sweep across a region. Deterministic root-mean-square error (RMSE),
the usual training and reporting metric, is dominated by smooth clear-sky periods
and hides exactly these events. Spatiotemporal GNNs are a popular tool for
distributed solar forecasting \citep{simeunovic2022}, but they almost universally
use a \emph{static, symmetric} graph, which ignores that cloud information has a direction and a
travel time set by the wind.

This paper asks a simple question: \emph{does making the graph advection-aware
actually help, and if so, when?} We build a controlled synthetic testbed in which
clouds are a Gaussian random field advected by a known, slowly varying wind, so
that a ground-truth CMV is available for every forecast origin. Against this
oracle we can cleanly separate three things that are usually entangled: the
benefit of the advection \emph{graph}, the benefit of an accurate motion
\emph{feature}, and the cost of \emph{estimating} the motion from data.

\paragraph{Contributions.}
\begin{enumerate}\itemsep2pt
\item A controlled decomposition of the advection-aware nowcasting problem
  (Sec.~\ref{sec:e1}--\ref{sec:decomp}): with a realistic CMV estimate the
  advection graph does not beat plain baselines; roughly half the oracle-CMV
  benefit is an accurate \emph{input feature}, not graph structure; and the
  benefit exists only inside a $v\cdot H \lesssim$ network-extent envelope.
\item A small \textbf{self-supervised cloud-motion estimator}
  (Sec.~\ref{sec:cmv}) that recovers the true wind to \SIrange{2}{4}{\degree}
  median angular error---$2$--$4\times$ better than the classical
  cross-correlation method---using only a multi-lag optical-flow reconstruction
  loss with an annealed kernel and a position-aware encoder.
\item A \textbf{frozen two-stage forecaster} (Sec.~\ref{sec:frozen}) that uses
  this estimator and closes $\sim$\SI{60}{\percent} of the oracle-CMV RMSE gap at
  moderate wind, with a clean operating envelope.
\item Negative results, reported plainly: the advection graph alone
  (Sec.~\ref{sec:e1}), coupling the motion estimate to the forecast loss
  (Sec.~\ref{sec:cmv}), and a spatially-coherent probabilistic head
  (Sec.~\ref{sec:e3}).
\end{enumerate}

\paragraph{Scope.} Every result here is on one synthetic simulator. The synthetic
``true wind'' is a generative parameter, so the CMV-recovery result in particular
must be read as a controlled sanity check, not a claim about real skies. We are
explicit about this in Sec.~\ref{sec:limits} and describe the real-data study it
calls for.

\section{Related work}

\paragraph{Spatiotemporal GNN forecasting.} Modelling a set of geographically
distributed series as signals on a graph and applying graph-convolutional or
graph-attention layers with a temporal encoder is now standard in traffic
forecasting---DCRNN \citep{li2018}, STGCN \citep{yu2018}, Graph WaveNet
\citep{wu2019}, ASTGCN \citep{guo2019}---and has been carried over to multi-site
PV power forecasting, e.g.\ the graph-convolutional LSTM/transformer models of
\citet{simeunovic2022}. The adjacency is almost always \emph{static}, built from
geographic distance or historical correlation; Graph WaveNet additionally learns
a global adjacency, which we include as a baseline. None of these encode cloud
advection explicitly.

\paragraph{Solar nowcasting and cloud motion.} Minutes-ahead irradiance
forecasting has a long history using sky imagers \citep{chow2011}, satellite
imagery, and cloud tracking; see \citet{yang2018} for a review. The dominant
non-image approach on a sparse ground network is to estimate a rigid
cloud-motion vector (CMV) by cross-correlating pairs of clear-sky-index series
and solving a least-squares problem for the translation velocity
\citep{bosch2013, lonij2013}, then ``advect the observed field forward''
\citep{nonnenmacher2014}. Our synthetic-field advection feature is the GNN
analogue of that heuristic, and our self-supervised estimator is an
optical-flow-style \citep{horn1981} objective specialised to a sparse sensor
layout.

\paragraph{Learned advection in weather nowcasting.} Precipitation nowcasting
has moved from ConvLSTM \citep{shi2015} to large advection-aware neural models
such as MetNet \citep{sonderby2020} and DGMR \citep{ravuri2021}, which learn
motion implicitly from dense radar rasters. Distributed solar sensing gives only
a sparse, irregular sample of the field, which is the regime we study.

\paragraph{Ramp forecasting and probabilistic scores.} Ramp events lack a single
definition; the swinging-door construction of \citet{florita2013} is a common
choice, which we adopt for our ramp-capture metrics. For the probabilistic head
we use strictly proper multivariate scores---the energy score \citep{gneiting2007}
and the variogram score \citep{scheuerer2015}. Clear-sky index uses the Haurwitz
model \citep{haurwitz1945}; the Ineichen--Perez model \citep{ineichen2002} is a
common alternative.

\section{Problem setup and synthetic testbed}
\label{sec:setup}
We forecast the clear-sky index $\kt = \mathrm{GHI}/\mathrm{GHI}_{\mathrm{cs}}$ at
$N$ fixed sites, $H$ steps ahead ($H=16$ at \SI{30}{\second}, i.e.\ 8 minutes).
Clouds are a periodic Gaussian random field of optical depth, translated by a
wind vector $\mathbf{v}(t)$ that performs a bounded random walk (a
\emph{steadiness} knob controls its variance); an \emph{evolving} option blends
keyframe fields so the pattern also grows and decays. Observed GHI is
clear-sky GHI \citep{haurwitz1945} times a transmittance function of the local
optical depth, plus noise. The simulator emits, per origin, the site series, the true
$\mathbf{v}(t)$ (the \emph{oracle} CMV), and a classical cross-correlation
estimate $\mathbf{v}_{\mathrm{est}}$ \citep{bosch2013}. Default layout: $N=49$ on a \SI{1}{km}
grid ($\sim$\SI{7}{km} across). We report test RMSE on $\kt$ and ramp-capture
metrics; means are over 2--3 seeds.

\section{Methods}
\label{sec:methods}
\paragraph{Advection-aware graph.} Given a motion vector $\mathbf{v}$, for each
site $j$ and horizon $h$ the arriving cloud is currently near
$\mathbf{p}_j - \mathbf{v}\,h\,\Delta t$. We connect $j@h$ to its $k$ nearest
sensors there with distance-weighted, per-$(j,h)$-normalised weights
$w \propto \exp(-d^2/\ell^2)$, $\ell$ the median nearest-neighbour spacing, and
pool their embeddings into an $(N,H,d)$ feature for the forecast head---the
GNN form of ``advect the field forward'' \citep{nonnenmacher2014}. An off-grid reliability gate
down-weights $(j,h)$ whose upwind point lands far from any sensor.

\paragraph{Forecaster.} A per-site GRU temporal encoder feeds $L$ heterogeneous
message-passing layers over $k$-NN and historical-correlation edge types, each
with its own message MLP and a gated residual update; the advection feature
enters at the head through a gated correction. Baselines: smart persistence,
per-site GRU, static-distance GCN, and a learned-adjacency GNN (a Graph WaveNet-style global
adjacency \citep{wu2019}). Ramp events use the swinging-door construction
\citep{florita2013}.

\paragraph{Self-supervised cloud-motion estimator.} A rigidly advecting field
obeys $\kt(\mathbf{x},t)=\kt(\mathbf{x}-\mathbf{v}L\Delta t,\,t-L)$. We train a
small position-aware encoder (per-site GRU $+$ a geometry-preserving DeepSets
pool) to output $\vhat$, minimising the residual of this identity, interpolated
from the sensor field, summed over several lags $L\in\{2,4,6,8,10\}$ (one $\vhat$
must explain them all):
\begin{equation}
  \mathcal{L}_{\mathrm{rec}}(\vhat) \;=\;
  \frac{1}{|\mathcal{L}|}\sum_{L}\Big\|\,\kt(\cdot,t)
  - \textstyle\sum_i \mathrm{softmax}_i\!\big(-\|\mathbf{p}_\cdot - \vhat L\Delta t
  - \mathbf{p}_i\|^2/\tau^2\big)\,\kt(\mathbf{p}_i, t-L)\,\Big\|^2 .
\end{equation}
The softmax width $\tau$ is annealed from $0.40$ to $0.10$ of the network
diameter (large $\tau$ gives useful gradients far from the optimum; small $\tau$
sharpens the estimate). An optional light MSE-to-reference term stands in for a
small amount of numerical-weather-prediction (NWP) wind on real data.

\paragraph{Two-stage forecaster.} We \emph{pre-train} the estimator, \emph{freeze}
it, and feed $\vhat$ into (a) the encoder's motion input channels and (b) the
advection graph. Freezing avoids a failure mode we observed when $\vhat$ is
trained jointly with the forecast loss: the loss drives $\vhat\to\mathbf{0}$
(Sec.~\ref{sec:cmv}).

\section{Experiments}

\subsection{Does an advection graph help? (E1)}
\label{sec:e1}
\begin{figure}[t]
  \centering
  \includegraphics[width=0.62\linewidth]{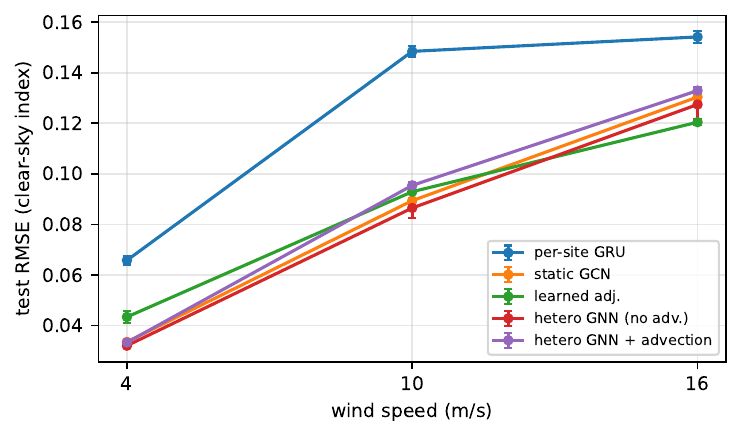}
  \caption{\textbf{E1.} Test RMSE vs.\ wind speed, estimated CMV, 2 seeds
    ($\pm$s.d.). The advection feature (top line) does not beat the heterogeneous
    GNN without it, nor the learned-adjacency baseline.}
  \label{fig:e1}
\end{figure}

With the classical CMV estimate, we sweep wind speed and compare the forecaster
with and without the advection feature against baselines (Fig.~\ref{fig:e1},
Table~\ref{tab:e1}). The heterogeneous GNN \emph{without} advection is the best or
tied-best model at low--moderate wind; a learned global adjacency wins at high
wind and under non-stationary/evolving clouds. Adding the advection feature on
top of the estimated CMV never helps and costs up to \SI{10}{\percent} RMSE at
\SI{10}{m/s}. On a dense grid every upwind sensor is already a graph neighbour, so
learned spatial mixing captures advection implicitly.

\begin{table}[t]
  \centering\small
  \caption{\textbf{E1} test RMSE (clear-sky index), estimated CMV, 2 seeds.
    Lower is better; best per column in \textbf{bold}.}
  \label{tab:e1}
  \begin{tabular}{lccc}
    \toprule
    & \multicolumn{3}{c}{wind speed (m/s)}\\ \cmidrule(lr){2-4}
    model & 4 & 10 & 16 \\
    \midrule
    per-site GRU            & 0.066 & 0.148 & 0.154 \\
    static-distance GCN     & 0.034 & 0.089 & 0.130 \\
    learned adjacency       & 0.043 & 0.093 & \textbf{0.120} \\
    hetero GNN (no adv.)    & \textbf{0.032} & \textbf{0.087} & 0.127 \\
    hetero GNN $+$ advection& 0.033 & 0.095 & 0.133 \\
    \bottomrule
  \end{tabular}
\end{table}

\subsection{The cloud-motion bottleneck (E2 and decomposition)}
\label{sec:decomp}
Replacing the estimated CMV with the \emph{oracle} flips the picture: with a
perfect wind vector the advection feature cuts RMSE by \SIrange{14}{17}{\percent}
and lifts ramp-down capture (CSI) by about 5 points (Table~\ref{tab:e2}). A
finer decomposition at \SI{10}{m/s} (2 seeds) shows \emph{where} that benefit
lives: no advection $0.087$; true wind in the graph only $0.082$; true wind in
the graph \emph{and} as an input feature $0.066$; oracle skyline $0.064$. Roughly
half of the oracle-CMV advantage is simply an accurate motion vector broadcast as
an input feature; the rest is graph structure.

\begin{table}[t]
  \centering\small
  \caption{\textbf{E2.} Advection feature value depends on CMV quality
    (test RMSE, 2 seeds).}
  \label{tab:e2}
  \begin{tabular}{llccc}
    \toprule
    wind & CMV & no adv. & $+$adv. & $\Delta$RMSE \\
    \midrule
    \SI{10}{m/s} & estimated & 0.086 & 0.096 & $+11\%$ \\
    \SI{10}{m/s} & \textbf{oracle} & 0.077 & \textbf{0.066} & $-14\%$ \\
    \SI{16}{m/s} & estimated & 0.133 & 0.128 & $-4\%$ \\
    \SI{16}{m/s} & \textbf{oracle} & 0.124 & \textbf{0.103} & $-17\%$ \\
    \bottomrule
  \end{tabular}
\end{table}

\subsection{Self-supervised cloud-motion estimation}
\label{sec:cmv}
\begin{figure}[t]
  \centering
  \includegraphics[width=0.8\linewidth]{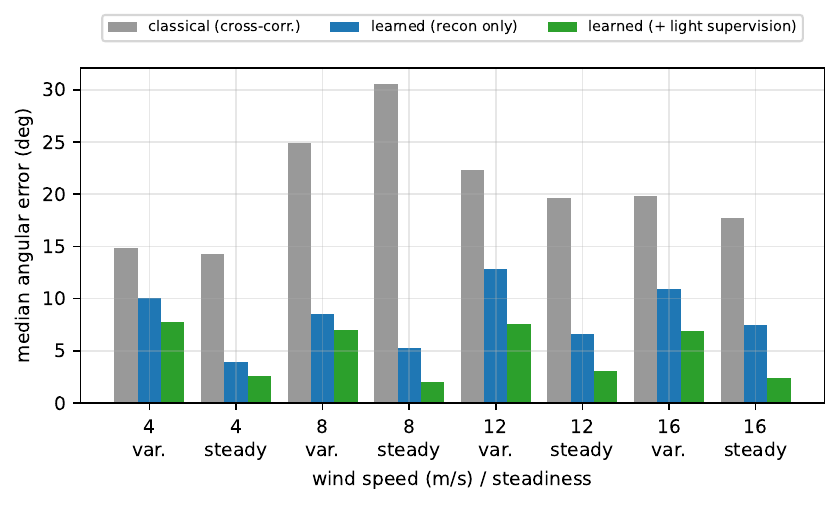}
  \caption{\textbf{Self-supervised cloud-motion estimation.} Median angular error
    of $\vhat$ vs.\ the true wind, 3 seeds. The learned estimator beats the
    classical cross-correlation method in every wind regime.}
  \label{fig:cmv}
\end{figure}

Trained standalone on $\mathcal{L}_{\mathrm{rec}}$, the estimator's median
angular error is \SIrange{2}{4}{\degree} in steady wind and
\SIrange{7}{13}{\degree} in variable wind---$2$--$4\times$ better than the
classical cross-correlation estimate (\SIrange{14}{31}{\degree}) in every regime
(Fig.~\ref{fig:cmv}, Table~\ref{tab:cmv}). A light supervision term
(weight $0.1$) tightens the hard cases to \SIrange{2}{8}{\degree}; it is not
required. Ablation: the position-aware encoder is what matters; kernel annealing
and the classical-estimate prior are minor.

\emph{Negative result.} An earlier design that reused the forecaster's
embeddings and trained $\vhat$ jointly with the forecast loss was unstable
(recovered cosine swung $0.3$--$0.99$ across module and loss variants); feeding
$\vhat$ to the head while coupled to the forecast loss collapses it to
$\vhat\to\mathbf{0}$. Decoupling (standalone encoder, frozen after
pre-training) removes this.

\begin{table}[t]
  \centering\small
  \caption{\textbf{CMV estimator} median angular error (deg), 3 seeds.}
  \label{tab:cmv}
  \begin{tabular}{lccc}
    \toprule
    regime & learned (recon) & learned ($+$ light sup.) & classical \\
    \midrule
    steady, 4--16 m/s   & 3.9--7.5 & \textbf{2.0--3.1} & 14--31 \\
    variable, 4--16 m/s & 8.5--12.8 & \textbf{6.9--7.7} & 15--25 \\
    \bottomrule
  \end{tabular}
\end{table}

\subsection{Frozen two-stage forecaster}
\label{sec:frozen}
\begin{figure}[t]
  \centering
  \includegraphics[width=0.66\linewidth]{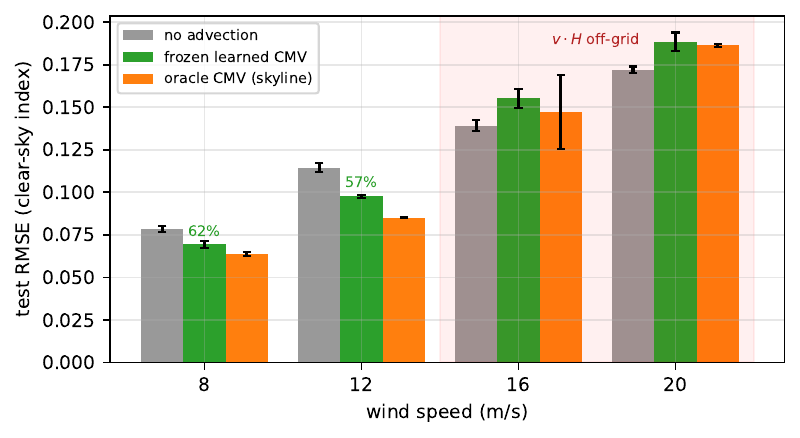}
  \caption{\textbf{Frozen two-stage forecaster}, 3 seeds ($\pm$s.d.). Green
    labels: fraction of the no-advection$\to$oracle gap closed. In the shaded
    region ($v\cdot H$ off-grid) even the oracle does not beat no advection.}
  \label{fig:frozen}
\end{figure}

Pre-training the estimator, freezing it, and feeding $\vhat$ to the forecaster
closes about \SI{60}{\percent} of the oracle-CMV RMSE gap at moderate wind
(\SIrange{8}{15}{\percent} RMSE reduction over no advection), tight across 3
seeds (Fig.~\ref{fig:frozen}, Table~\ref{tab:frozen}). At $\geq$\SI{16}{m/s}
advection does not help \emph{even with the oracle}: the advective displacement
over the horizon (\SIrange{7.7}{9.6}{km}) exceeds the $\sim$\SI{7}{km} network,
so there is no on-grid upwind information. The off-grid gate prevents a
catastrophic regression but cannot manufacture signal. \textbf{Operating
envelope: advection nowcasting helps iff $v\cdot H$ fits within the network.}

\begin{table}[t]
  \centering\small
  \caption{\textbf{Frozen two-stage} test RMSE (clear-sky index), 3 seeds.}
  \label{tab:frozen}
  \begin{tabular}{lcccc}
    \toprule
    wind (m/s) & no adv. & frozen CMV & oracle & CMV err. (deg) \\
    \midrule
    8  & 0.078 & \textbf{0.069} & 0.064 & 2.8 \\
    12 & 0.114 & \textbf{0.098} & 0.085 & 3.9 \\
    16 & 0.139 & 0.155 & 0.147 & 2.9 \\
    20 & 0.172 & 0.188 & 0.186 & 2.4 \\
    \bottomrule
  \end{tabular}
\end{table}

\subsection{Spatially-coherent probabilistic head (negative)}
\label{sec:e3}
We tried a joint low-rank-plus-diagonal Gaussian head across sites (for
fleet-aggregate ramp risk), trained by exact Woodbury NLL, versus a diagonal
head. It was marginally \emph{worse} on every metric, including the energy
\citep{gneiting2007} and variogram \citep{scheuerer2015} scores it is meant to
improve (e.g.\ at \SI{16}{m/s}: variogram
$52.6$ vs.\ $51.0$; CRPS $0.065$ vs.\ $0.063$). After conditioning on the graph
features the site residuals appear to carry little extra dependence; we report
this as a negative result pending a dedicated study.

\section{Discussion}

\paragraph{When to use what.} The results give a concrete rule. If $v\cdot H$
fits inside the sensor network and the wind is reasonably steady, an
advection-aware feature driven by a good CMV estimate is worth
\SIrange{8}{15}{\percent} RMSE. Otherwise---fast wind, very sparse or irregular
networks, strongly non-stationary clouds---a learned global adjacency is the
better tool, and the advection feature is at best neutral (with the gate) and at
worst a \SI{10}{\percent} regression (without it).

\paragraph{The estimator is the reusable piece.} Independent of the forecasting
question, the self-supervised CMV estimator is a drop-in replacement for
cross-correlation on any bare sensor network, several times more accurate here,
and it needs no labels.

\paragraph{Limitations.}
\label{sec:limits}
All experiments use one synthetic simulator, and its ``true wind'' is a
generative parameter---so the CMV-recovery numbers are a controlled sanity check,
not a measurement on real skies, and the oracle gap that Sec.~\ref{sec:frozen}
closes is defined by that simulator. Real clouds have multiple layers, shear,
growth and decay, and no single motion vector. The necessary next step is
validation on a real distributed network---e.g.\ the NREL Oahu Solar Measurement
Grid \citep{osmg} (17 sensors, \SI{1}{km}, \SI{1}{Hz}) for the forecaster, and a sky-image
dataset with optical-flow motion for the estimator. Our released code includes
loaders and configs for these; only the raw data acquisition remains.

\section{Conclusion}
On a controlled testbed, an advection-aware graph feature for distributed solar
ramp nowcasting helps only within a $v\cdot H \lesssim \text{network-extent}$
envelope, and about half of its value is an accurate motion \emph{feature} rather
than graph structure. A small self-supervised estimator supplies that motion
several times more accurately than the classical method, and a frozen two-stage
forecaster built on it closes $\sim$\SI{60}{\percent} of the oracle-CMV gap at
moderate wind. We release the simulator, models, metrics, and experiment scripts.

\paragraph{Reproducibility.} The simulator, models, metrics, baselines, and
experiment scripts are available at
\url{https://github.com/appsofa-com/gnn-solar-energy} (branch
\texttt{solar-ramp-nowcasting}); every table and figure is regenerated from
committed result files. \texttt{pytest} covers the simulator, graph
construction, metrics, and model shapes (40 tests).

\bibliographystyle{plainnat}
\bibliography{refs}

\end{document}